# Learning to track on-the-fly using a particle filter with annealed-weighted QPSO modeled after a singular Dirac delta potential

**Saptarshi Sengupta**
S. Sengupta is with the Department of Electrical Engineering and Computer Science, Vanderbilt University, Nashville TN 37212, USA. (Corresponding author, Email: saptarshi.sengupta@vanderbilt.edu; sengupta.eecs@yahoo.com; Tel: +1 (615)-678-3419)

**Richard Alan Peters II**
R.A. Peters II is with the Department of Electrical Engineering and Computer Science, Vanderbilt University, Nashville TN 37212, USA. (Email: Alan.Peters@Vanderbilt.Edu.

*Abstract* – This paper proposes an evolutionary Particle Filter with a memory guided proposal step size update and an improved, fully-connected Quantum-behaved Particle Swarm Optimization (QPSO) resampling scheme for visual tracking applications. The proposal update step uses importance weights proportional to velocities encountered in recent memory to limit the swarm movement within probable regions of interest. The QPSO resampling scheme uses a fitness weighted mean best update to bias the swarm towards the fittest section of particles while also employing a simulated annealing operator to avoid subpar fine tune during latter course of iterations. By moving particles closer to high likelihood landscapes of the posterior distribution using such constructs, the sample impoverishment problem that plagues the Particle Filter is mitigated to a great extent. Experimental results using benchmark sequences imply that the proposed method outperforms competitive candidate trackers such as the Particle Filter and the traditional Particle Swarm Optimization based Particle Filter on a suite of tracker performance indices.

## 1. Introduction

Visual Object Tracking is an active research area within the Computer Vision community and has been rigorously studied due to its relevance in achieving key practical functionalities in today's increasingly complex cyber-physical world. Some of the more well-known applications include real-time video surveillance and security systems, smart traffic monitoring and autonomous vehicle navigation. While object trackers aim to identify distinguishing features of targets across multiple frames of interest in sequential images, several challenging issues arise that pose as potential failure modes. Varying environmental and behavioral conditions such as complex object motion, partial or complete occlusion of the region of interest, changes in illumination and scale, injection of noise etc. lead to inefficient and at many times failed tracking. Constrained optimization approaches in mitigating tracking failures have demonstrated notable success. The existing methods use either deterministic [1-3] or stochastic approaches [4-11]. Deterministic approaches typically employ gradient descent search in order to minimize a cost function and obtain parametric estimates. One such example that has been extensively used is the Snakes model introduced by Kass et al [1]. Hager and Belhumer defined the cost function as the sum of squared deviations of candidate solutions from the ground truth [2] whereas Comaniciu minimized the cost difference between two color histograms by using the Mean Shift Algorithm [3]. Deterministic approaches are computationally less expensive, however they are susceptible to getting trapped in local optima. Stochastic approaches involve probabilistic operators and better estimate parameters by intelligently querying the multidimensional search space for the global optima, with the tradeoff being computational load. Several approaches have been proposed in [4-11] which effect better performance compared to their deterministic counterparts but the curse of dimensionality remains for high dimensional problems. Due to the dynamic nature of the environment, a unified object tracking scheme is very difficult to accomplish. Particle Filters are recursive implementations of Monte Carlo methods and are ideal for analyzing highly non-linear, non-Gaussian state estimation problems where classical Kalman Filter based approaches fail [12]. The generic Particle Filter suffers from the degeneracy and Sampling Importance Resampling (SIR) induced particle impoverishment problem, leading to proposed enhancements in the sampling stage as in [13-14].

This paper incorporates a memory guided motion model and a hybrid QPSO resampling scheme using annealing and weighted mean best operators (Annealed Weighted QPSO-AWQPSO) to effectively recast particles to the higher likelihood regions in the posterior probability landscape. The methodology is tested out on two benchmark problems containing a set of environmental test conditions. Performance metrics like Root Mean Square Error (RMSE), Number of Frames Successfully Tracked, Tracking Precision versus Centre Error Threshold, Recall versus Overlap Threshold and Frames per Second (FPS) are analyzed over batches of computations. Such statistical analyses suggest performance improvements using the proposed method in comparison to the PSO Resampling inspired Particle Filter (PSO-PF) as well as the standard Particle Filter (PF).

The rest of the paper is summarized as follows: Section 2 reviews related work at the intersection of Evolutionary Computation and particle resampling in Particle Filters, Section 3 outlines the resampling techniques used and Section 4 details the proposed approach. Section 5 lists the tracking quality indices used in the model followed by Section 6 which elaborates on the experimental conditions and results on benchmark problems. Section 7 provides an analysis of the results obtained and Section 8 concludes the paper with possible directions for future work.

## 2. Sample impoverishment in particle filters and related work

A Bayesian inference approach to the object tracking problem involves dynamic state transition through time using a System Model and state measurement through an Observation Model. A Markovian system model in this regard can be formulated as:

$$X_k = f(X_{k-1}, v_k) \leftrightarrow p(X_k | X_{k-1}) \tag{1}$$

The observation model can be expressed as:

$$O_k = h(X_k, \eta_k) \leftrightarrow p(O_k | X_k) \tag{2}$$

The sequences $\{X_k, k \in I^+\}$ along with $\{O_k, k \in N\}$ denote the target states and the measurement set of the state sequence in frame $k$. $v_k$ and $\eta_k$ are mutually independent system noise and measurement noise. The central goal of a particle filter is to find an approximation of the posterior probability distribution $p(O_k/X_k)$ using a set of weighted samples drawn from a proposal distribution with an associated particle rank defined by a one to one correspondence between high posterior likelihood and large weight. The weights $\kappa$ are generally computed using the following proportionality relation:

$$\kappa_k^i \propto \kappa_{k-1}^i \frac{p(O_k | X_k^i) p(X_k^i | X_{k-1}^i)}{p(X_k | X_{k-1}^i, O_k)} \tag{3}$$

The posterior distribution is then updated as:

$$p(X_k | O_k) = \sum_{i=1}^{N} \kappa_k^i \, \delta(X_k - X_k^i) \tag{4}$$

where $p(O_k/X_k)$ is the likelihood and $\delta(.)$ is the Dirac-delta function. It is fundamentally important to generate a proposal distribution such that the sampled particles belong to the region of significant likelihood of the posterior. Given that particle filters run into sample degeneracy issues [15] because a large fraction of particles have negligibly small weights after only a few iterations, Sampling Importance Resampling (SIR) based probabilistic selection of particles have been widely adopted as a solution [16]. In the resampling step, particles having small weights have low chances of being propagated to the next iteration. A key disability of PF-SIR in effectively addressing the *Sample Degeneracy Problem* lies in loss of particle diversity over the course of iterations. This leads to the *Sample Impoverishment Problem* [17] as the resampled particle set does not accurately reflect the underlying statistical properties of the original particle set. As the number of effective particles decreases, the collective information carried by them also declines resulting in suboptimal object representations. The number of effective particles $N_{eff}$ can be expressed as:

$$N_{eff} = \frac{1}{\sum_{i=1}^{N} (\kappa_k^i)^2} \tag{5}$$

The *Sample Impoverishment Problem* has attracted several mitigation strategies that make use of prior knowledge processing or multi-layered sampling. Partitioned Sampling [17], Annealed Importance Sampling [19] and Kernel Particle Filters [13] are some of the commonly used techniques in this regard. The Auxiliary Particle Filter by *Pitt* and *Shephard*, 1999 [20] samples particles corresponding to points mapped to an importance density with high conditional likelihood. Some researchers have proposed moving particles of lower importance towards regions of higher posterior likelihood. For example, the Kernel Particle Filter accomplishes this particular objective, however it uses a deterministic search and requires a continuous probability distribution, among other things.

In recent years, the use of Particle Swarm Optimization [21] in non-differentiable and ill-structured multidimensional problems has gained popularity due to co-operative exchange of social and cognitive information among swarm members and the relatively low cost of individual particle fitness computation. While it yields promising results for non-differentiable cost functions, it is also limited in its ability to converge to the global best (Van den Bergh, 2001) [23] as per the convergence criteria put forward by Solis and Wet [24]. Numerous updates to the canonical PSO put forward by Clerc and Kennedy [25] have been made possible by factoring in different initialization conditions, position and velocity updates and hybridization [22] [25-27] [31]. Of these, Quantum-behaved Particle Swarm Optimization (QPSO) [26-30] is a particularly attractive choice as its convergence to optima is theoretically guaranteed [31]. Promising results using QPSO-inspired Particle Filters in several tracking datasets have been reported by Sun et al (2015) [7] and by Hu, Fang and Ding (2016) [8].

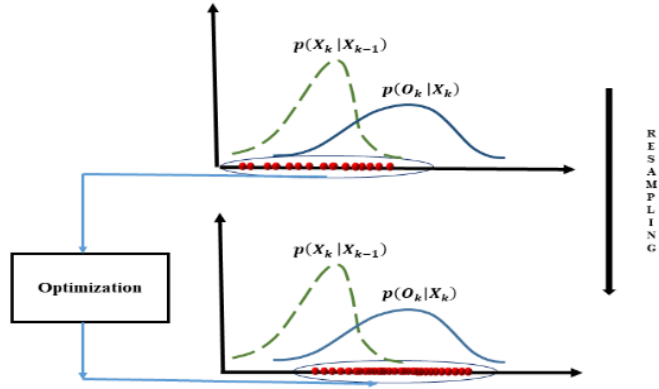

**Fig. 1**. Particle redistribution towards regions of high likelihood.

## 3. Outline of metaheuristics used

*3.1. Particle swarm optimization (PSO)*

PSO [21-22] is one of many nature-inspired metaheuristics in the broad category of Swarm Intelligence and draws motivation from social co-operation among bird flocks or fish schools. Each particle in PSO is a candidate solution representing a point in a *d-dimensional* search space. The particles mimic the behavior exhibited by a swarm of birds flocking in a multidimensional search space by updating their position coordinates and velocity using information of personal best position so far (*cognitive operator - pbest)* and global best (*social operator - gbest*). An iterative process of movement dependent on social co-operation guides the swarm towards the global optima. The position and velocity equations in basic PSO are as follows:

$$v_i^{t+1} = \omega\, v_i^t + C_1\, r_1(pBest_i - X_i^t) + C_2 r_2(gBest - X_i^t) \qquad (6)$$

$$X_i^{t+1} = X_i^t + v_i^{t+1} \qquad (7)$$

$C_1$ and $C_2$ are cognition and social acceleration constants and $r_1$ and $r_2$ are random numbers between 0 and 1 drawn from a uniform distribution. $X_i^{t+1}$, $v_i^{t+1}$ represent the position and velocity of the $i_{\text{th}}$ *d*-dimensional particle respectively at the end of the *t*-th iteration whereas $pBest$ and $gBest$ are the personal and global best positions. Term 1 in the R.H.S of eq. (6) represents inertia of the swarm and can be adjusted by tuning $\omega$ while the next two terms perturb noise in the direction of the individual and population best. The fitness *f* is updated in the following manner for a cost minimization objective:

$$f(x_i^t) < f(pBest_i)) \Rightarrow pBest_i = x_i^t \qquad (8)$$

$$f(x_i^t) \geq f(pBest_i)) \Rightarrow pBest_i = pBest_i \qquad (9)$$

**Algorithm 1.** *Particle Swarm Optimization*

```
1: for each particle xi
2:     initialize position and velocity
3: end for
4: do
5:   for each particle xi
6:       Calculate particle fitness fi
7:       if fi is better than individual best (pBest)
8:           Set fi as the new pBest
9:       end if
10:  end for
11: Set best among pBest as the global best (gBest)
12: for each particle
13:      Calc. particle velocity acc. to eq. (6)
14:      Update particle position acc. to eq. (7)
15: end
16: while max. iter or convergence criterion  not met
```

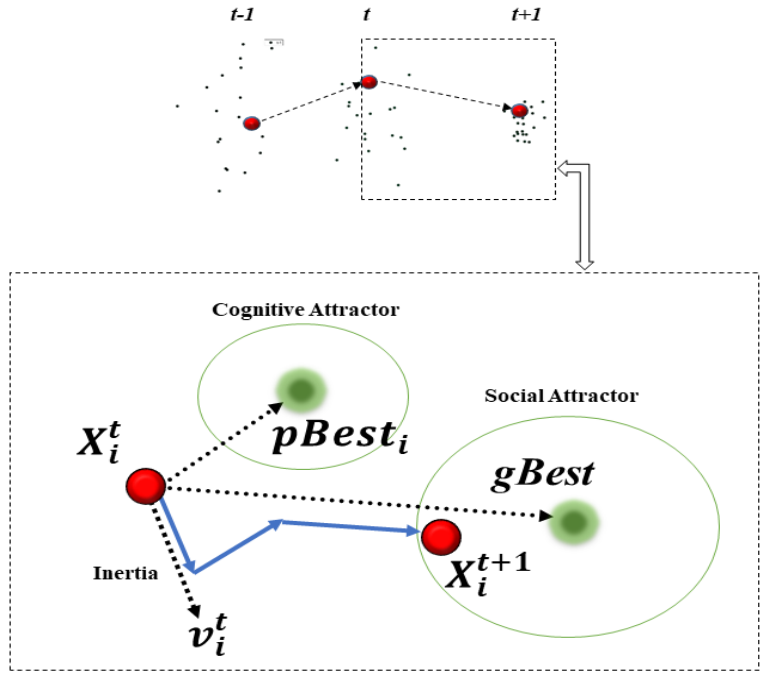

**Fig. 2.** Particle movement mechanics using PSO.

*3.2. Quantum-behaved particle swarm optimization (QPSO)*

Trajectory analysis in [33] proved that the convergence of PSO necessitates the convergence of each particle to its local attractor $p_i^t = (p_{i1}^t, p_{i2}^t, p_{i3}^t, \ldots p_{id}^t)$ and in the process the current position ($X_i^t$), the personal best (*pBest*) and the global best (*gBest*) approach the same value. In Quantum-behaved Particle Swarm Optimization, the state of a particle is formally characterized by a wave function $\psi$ with $|\psi|^2$ representing the probability density function of its position. Using Monte Carlo recursion, the QPSO coordinate update equation reduces to:

$$X_{ij}^{t+1} = p_{ij}^t \pm \left(\frac{L_{ij}^t}{2}\right) \ln\left(\frac{1}{u_{ij}^t}\right) \qquad (10)$$

$u_{ij}^t \sim U(0,1)$ is a uniformly distributed random number and the local attractor $p_{ij}^t$ can be formulated as:

$$p_{ij}^t = \frac{C_1 rand(0,1)_{ij}^t pbest_{ij}^t + C_2 rand(0,1)_{ij}^t gBest_{ij}^t}{C_1 rand()_{ij}^t + C_2 rand()_{ij}^t} \qquad (11)$$

*rand(0,1)* generates different random numbers for pairing with cognitive and social operators. Further simplification results in the following widely used form:

$$p_{ij}^t = \Phi_{ij}^t pBest_{ij}^t + (1-\Phi_{ij}^t) gBest_{ij}^t \qquad (12)$$

where $\Phi_{ij}^t \sim U(0,1)$ is a generated random number distributed uniformly.

The parameter $L_{ij}^t$ is the characteristic length of the underlying wave function and is evaluated as:

$$L_{ij}^t = 2\beta \left| p_i^t - X_{ij}^t \right| \qquad (13)$$

The contraction-expansion co-efficient $\beta$ is tuned to maintain the balance between exploration and exploitation. The complete position update equation is thus given by:

$$X_{ij}^{t+1} = p_{ij}^t \pm \beta \left| p_i^t - X_{ij}^t \right| \ln\left(\frac{1}{u_{ij}^t}\right) \qquad (14)$$

$L_{ij}^t$ controls the accuracy and convergence speed of QPSO. The *"Mainstream Thought"* or *Mean Best*, introduced in [26] is the mean of all *pBest* positions of the particles.

$$mbest^t = (mbest_1^t, mbest_2^t, \ldots, mbest_d^t) \qquad (15)$$
$$= \left[ \frac{1}{M}\sum_{i=1}^{M} p_{i1}^t, \frac{1}{M}\sum_{i=1}^{M} p_{i2}^t, \ldots, \frac{1}{M}\sum_{i=1}^{M} p_{id}^t \right]$$

An alternate way of writing the position update equation is adopted by re-expressing $L_{ij}^t$:

$$L_{ij}^t = 2\beta \left| mbest_j^t - X_{ij}^t \right| \qquad (16)$$

This yields the final form of the popular mainstream thought based position update equation of the QPSO algorithm.

$$X_{ij}^{t+1} = p_{ij}^t \pm \beta \left| mbest_j^t - X_{ij}^t \right| \ln\left(\frac{1}{u_{ij}^t}\right) \qquad (17)$$

The second term in the RHS of (17) is additive when a generated random number is less than 0.5 and vice-versa.

**Algorithm 2.** *Quantum-behaved PSO*

```
1:  for each particle x_i
2:      initialize position
3:  end for
4:  do
5:     Compute mean best position using eq. (15)
6:      for each particle x_i
7:          for each dimension j
8:              Calculate local attractor using eq. (12)
9:              if rand(0,1)<0.5
10:                 Update pos. using eq. (15) with '+'
11:             else Update pos. using eq. (15) with '-'
12:             end if
13:         end for
14:         Evaluate fitness function
15:         Update pBest according to eq. (8) and (9)
16:     end for
17:    Set best among pBest as the global best (gBest)
18: while max. iter or convergence criterion  not met
```

## 4. Annealed-weighted QPSO for visual tracking

*4.1. Particle propagation using AWQPSO*

The uniform weighting scheme in the *Mean Best* calculation in eq. (15) is not an optimum choice as particles of varying fitness values contribute equally to it. Thus, in alignment with predator-prey population models where the fitter of the two survives to pass on their genes, the mean best update is recomputed by assigning a set of variable weights with the particles. Each particle is associated with a weight in proportion to its fitness value thereby making it favorable for the fittest particle to contribute most to the mean best update [31]. The *mbest* calculation thus changes to:

$$mbest^t = (mbest_1^t, mbest_2^t, \ldots, mbest_d^t) \quad (18)$$
$$= \left[\frac{1}{M}\sum_{i=1}^{M} \tau_{i1}^t p_{i1}^t, \frac{1}{M}\sum_{i=1}^{M} \tau_{i2}^t p_{i2}^t, \ldots, \frac{1}{M}\sum_{i=1}^{M} \tau_{id}^t p_{id}^t\right]$$

where $\tau_{ij}^t$ is the *j*-th dimensional weight of the *i*-th particle in iteration *t*. The standard QPSO suffers from unsatisfactory fine tune during the latter part of the search process [33] and the fitness update scheme rejects particles whose likelihood values are worse than the personal best. However, these particles may evolve over iterations to guide the swarm towards the globally optimum mode and disregarding them from the start of the search process may effectively reduce the diversity of the swarm. Thus, the fitness update scheme is replaced by an exponential acceptance score where the probability of accepting a particular particle is given by the Metropolis criterion [34]:

$$\theta = \begin{cases} 1, & \text{if } \Delta f < 0 \\ exp\left(-\frac{\Delta f}{T_t}\right), & \text{otherwise} \end{cases} \quad (19)$$

$\Delta f$ is the difference in fitness from previous iteration, $\theta$ is probability that the current particle is accepted and $T_t$ is the annealing temperature in iteration *t*. A suitable cooling schedule is adopted with an initial high value of $T_0$:

$$T_t = T_0(e^{-t}) \quad (20)$$

The value of the contraction-expansion factor $\beta$ is decreased linearly from 0.9 to 0.5 over the iteration count to facilitate exploitation in the latter part of the search:

$$\beta = (0.9 - 0.5)\left[\frac{(t_{max} - t_{current})}{t_{max}}\right] + 0.5 \quad (21)$$

**Algorithm 3.** *Annealed Weighted QPSO*
1: **for** each particle $x_i$
2:    initialize position
3: **end**
4: **do**
5:   Compute mean best position using eq. (18)
6:   **for** each particle $x_i$
7:     **for** each dimension j
8:       Calculate local attractor using eq. (12)
9:       **if** rand(0,1)<0.5
10:         Update pos. using eq. (17) with '+'
11:       **else** Update pos. using eq.(17) with '-'
12:       **end if**
13:     **end for**
14:   Accept new solution according to eq. (19)
15:   Update pBest according to. eq. (8) and (9)
16:   **end for**
17:   Set best among pBest as the global best (gBest)
18: **while** max. iter or convergence criterion not met

*4.2. Motion model and target observation*

The dynamic state update stage of the filter makes use of a weight normalized velocity looking back three steps in memory. A Gaussian distribution $X_{k+1} \sim N(X_k, \Sigma_M)$ is used to spread particles around the current state which results in the following motion model with the importance weight vector $\lambda$ sorted in ascending order of values. $\Sigma_M$ is the covariance matrix of the distribution, $v_f$ is the adaptive step size update, $\Omega$ is a uniform random number in [-1,1] and $v_g$ is the velocity of the $g$-th frame.

$$X_{k+1} = X_k + \Omega v_f \tag{22}$$

$$\lambda = sort\left(\left\{\frac{v_g}{\sum_{e=k-2}^{k} v_e}\right\}_{k-2}^{k}, ascending\right) \tag{23}$$

$$v_f = 2\sum_{a=0}^{2} \lambda\{a+1\}.\{v_{k-a}\} \tag{24}$$

Now, it is known to all that a good observation model is critical to implementing an efficient tracker. However, in practice varying conditions necessitate the use of specific feature descriptors for different tracking scenarios. In this work, the appearance of the targeted object is modeled using a Gaussian fitness function as:

$$f(C, \Sigma) = \left(\frac{1}{2\pi^{n/2}|\Sigma|^{1/2}}\right) \exp\left(-\frac{\Delta^2}{2}\right) \tag{25}$$

$\Delta = \sqrt{(C - C_{GT})^T \Sigma^{-1}(C - C_{GT})}$ is the *Mahalanobis distance* of the observable $C$ with respect to the goal state $C_{GT}$ given covariance $\Sigma$. Here, color cue is used as the feature descriptor to construct likelihood scores because of its simplicity in implementation while providing invariance to translational and rotational change, as well as scale change and partial occlusion. The Euclidean distance between $i$-th of $N$ particles and the manually annotated ground truth for the $k$-th frame is used in subsequent center error estimation and is given by the following equation:

$$d_i^k = \sqrt{(X_{GT} - X_i^k)^2} \quad \forall i \; in \; I^+ \in [1, N] \tag{26}$$

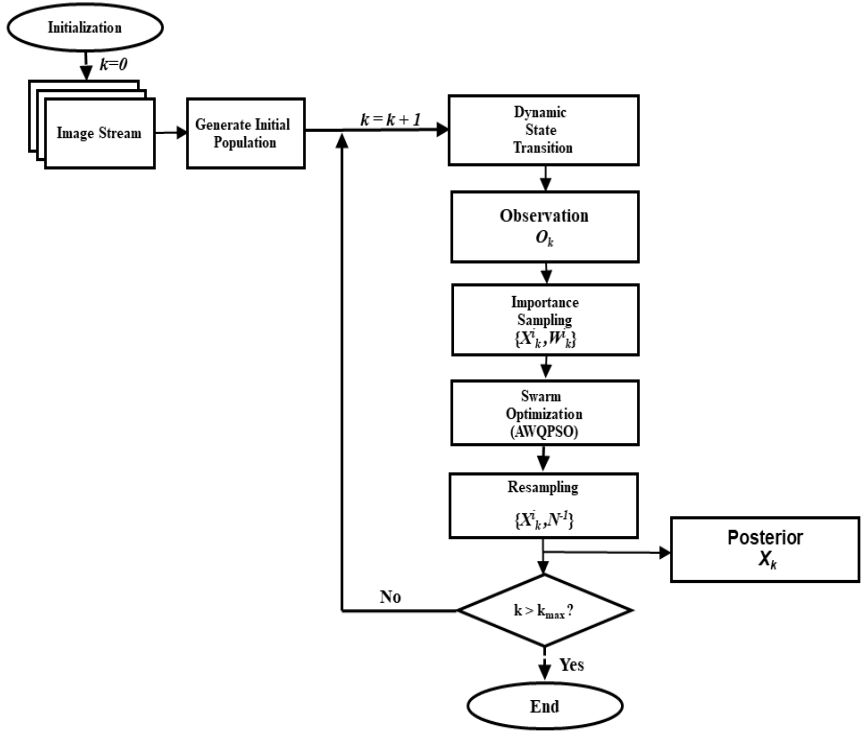

**Fig. 3.** Flowchart of the AWQPSO tracking model.

## 5. Tracking performance indices

A quantitative characterization of tracker performance has been made using precision and recall evaluated over the test sequences. Precision, in the context of visual tracking can be defined as the ratio of the number of frames over the total having a center to swarm deviation less than a preset threshold. Recall, on the other hand is the ratio of number of frames over the total that pass a tracker to ground truth bounding box overlap score greater than a preset threshold. In more formal terms, these are expressed as:

$$Precision = \frac{Frames_{RMSE < Threshold}}{Total\ Frames} = \frac{TP}{TP+FP} \tag{27}$$

$$Recall = \frac{Frames_{Overlap\ Score > Threshold}}{Total\ Frames} = \frac{TP}{TP+FN} \tag{28}$$

The Overlap Score is computed as $\left(\frac{BB_{Tracker} \cap BB_{Ground\ Truth}}{BB_{Tracker} \cup BB_{Ground\ Truth}}\right)$. TP, FP and FN are true positives, false positives and false negatives, respectively and BB denotes the Bounding Box.

$$RMSE = \frac{\sum_{k=1}^{Frames}\left[\frac{\sqrt{\sum_{z=1}^{N}\left\{(X_{z,x}-X_{GT,x})_k^2 + (X_{z,y}-X_{GT,y})_k^2\right\}}}{particles}\right]}{Frames} \tag{29}$$

$$FPS = \frac{Total\ Number\ of\ Frames\ Executed}{Processing\ Time\ Required\ In\ Seconds} \tag{30}$$

## 6. Experiments and results

### 6.1. Experimental setup

To evaluate the performance of the AWQPSO tracker, three competitive tracking algorithms viz. PF (described in Section 2), PSO – PF (described in Section 3) and AWQPSO-PF (described in Section 4) have been considered. A comparative analysis of computational load and error margins are calculated using the same observation model for all. Two different video sequences acquired at 25 fps are taken. The first one is the dataset *OneStopNoEnter2cor.mpg* from the EC Funded CAVIAR project/IST 2001 37540 [35]. The Corridor Views of the Lisbon Sequence from the CAVIAR Project are considered. These sequences are shot in a shopping mall using a surveillance camera and variations include scale change, different lighting conditions, nearby moving object (particle hijacking problem) and partial occlusion. The second sequence is *aerobatics_1.avi* from the *Aircraft Tracking Database-Open Remote Sensing* [36] which introduces scale change, camera movement, abrupt motion and specular reflection into the observation.

The values of the cognitive and social learning constants *C1* and *C2* in Table 2 are both set to 2.05 as these are empirically found to be the optimal pair. The inertial constant $\omega$ in PSO is set to 0.5 after testing a linear time varying inertia weight (TVIW) as well as in increments of 0.1 between 0.1 and 0.9 for PSO which results in a fine balance between exploration and exploitation. The contraction-expansion factor $\beta$ in AWQPSO is reduced linearly with the number of iterations to explore the search space more in initial iterations and hone in on potential solution regions towards the latter iterations. The population size in all test cases are taken to be 300 to allow for reasonably on-target behavior across all frames for each algorithm, exceeding which the time complexity increases with negligible change in the number of off-target frames. A sufficiently large fitness score computed with respect to the goal state or a maximum iteration count of 50 are kept as the termination criterion for all in-frame optimization using the algorithms.

The methodologies discussed so far are implemented on MATLAB R2016a using an Intel(R) Core(TM) i7-5500U CPU @ 2.40GHz with 8GB RAM and the performances over 30 trials are analyzed. No use of Graphics Processing Units (GPUs) have been made during the experiments.

**Table 1.**

**List of Implementation Terms and Parameters for the Metaheuristic Algorithms.**

| Term | Discussion |
| --- | --- |
| **Some General Terms** | |
| **Population (X)** | The collection or *'swarm'* of agents employed in the search space |
| **Fitness Function (f)** | A measure of convergence efficiency |
| **Current Iteration** | The ongoing iteration among a batch of dependent/independent runs |
| **Maximum Iteration Count** | The maximum number of times runs are to be performed |
| **Particle Filter** | |
| **Population (X)** | Collection of agents approximating states of target under consideration |
| **Proposal** | Initial guess of possible target states given some/no apriori knowledge |
| **Observation** | Sensed states of the target after the prediction stage is complete |
| **Importance Weights (κ)** | A high posterior likelihood implies a large weight |
| **Effective Sample Size (ESS)** | Low value of Effective Sample Size implies necessity of resampling |
| **Particle Swarm Optimization** | |
| **Position (X)** | Position values of individual swarm members employed in a multidimensional search space |
| **Velocity (v)** | Velocity values of individual swarm members |
| **Cognitive Accl. Coefficient (C1)** | Empirically found scale factor of pBest attractor |
| **Social Accl. Co-efficient (C2)** | Empirically found scale factor of gBest attractor |
| **Personal Best (pBest)** | Position corresponding to historically best fitness for a swarm member |
| **Global Best (gBest)** | Position corresponding to best fitness over history for swarm members |
| **Inertia Weight Co-efficient (ω)** | Facilitates and modulates exploration in the search space |
| **Cognitive Random Perturbation ($r_1$)** | Random noise injector in the Personal Best attractor |
| **Social Random Perturbation ($r_2$)** | Random noise injector in the Global Best attractor |
| **Quantum-behaved Particle Swarm Optimization** | |
| **Local Attractor** | Set of local attractors in all dimensions |
| **Characteristic Length** | Measure of scales on which significant variations occur |
| **Contraction-Expansion Param. (β)** | Scale factor influencing the convergence speed of QPSO |
| **Mean Best** | Mean of all personal bests across all particles, akin to leader election in the biological world |
| **Annealed Weighted Quantum-behaved Particle Swarm Optimization** | |
| **Weighted Mean Best** | Fitness weighted mean of all personal bests across all particles |
| **Metropolis Criterion** | Criterion facilitating inclusion of worse performing particles in the solution pool to preserve diversity of the swarm |
| **Annealing Temperature** | Temperature of the system in a particular iteration in the simulated annealing process [33] |
| **Initial Annealing Temperature** | Initial temperature of the system in the simulated annealing process |
| **Contraction Expansion Parameter (β)** | Linearly decreasing factor influencing convergence speed of QPSO |

**Table 2**
Parameter selection for the tracking algorithms.

| Parameter | Population | C1 | C2 | ω | β | $t_{max}$ | $T_t$ |
| --- | --- | --- | --- | --- | --- | --- | --- |
| **Value** | 300 | 2.05 | 2.05 | 0.5 | $(0.9-0.5)[(t_{max}-t_{current})/t_{max}]+0.5$ | 50 | 100 |

*6.2. Results for Benchmark Problem 1: OneStopNoEnter2cor*

| Frame | PF | PSO-PF | AWQPSO-PF |
|---|---|---|---|
| 805 | 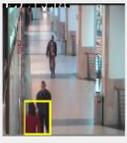 | 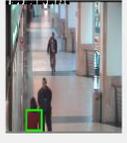 | 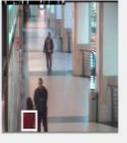 |
| 897 | 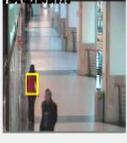 | 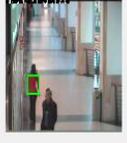 | 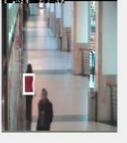 |
| 966 | 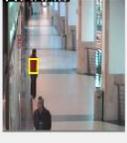 | 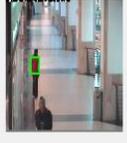 | 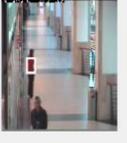 |
| 1035 | 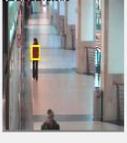 | 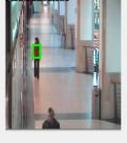 | 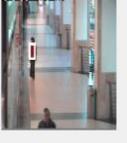 |
| 1081 | 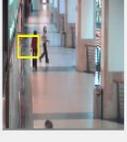 | 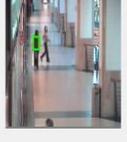 | 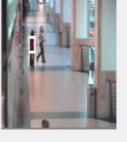 |

**Fig. 4.** Tracking results for OneStopNoEnter2cor.

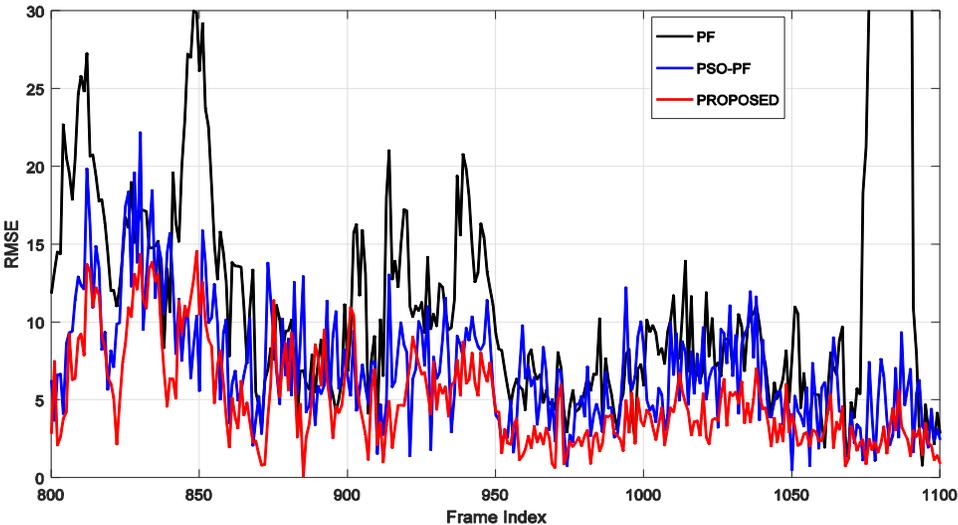

**Fig. 5.** Evolution of RMSE for OneStopNoEnter2cor (301 frames).

**Table 3**
Performance comparison of the three trackers for OneStopNoEnter2cor.

| Dataset | Algorithm | FPS | Lost Targets | |
|---|---|---|---|---|
| | | | CET=20 | CET=30 |
| OneStopNoEnter2cor | PF | **17.23±0.3058** | 40/301 | 28/301 |
| | PSO-PF | 6.71±0.7285 | 4/301 | **0/301** |
| | AWQPSO-PF | 8.69±0.7044 | **0/301** | **0/301** |

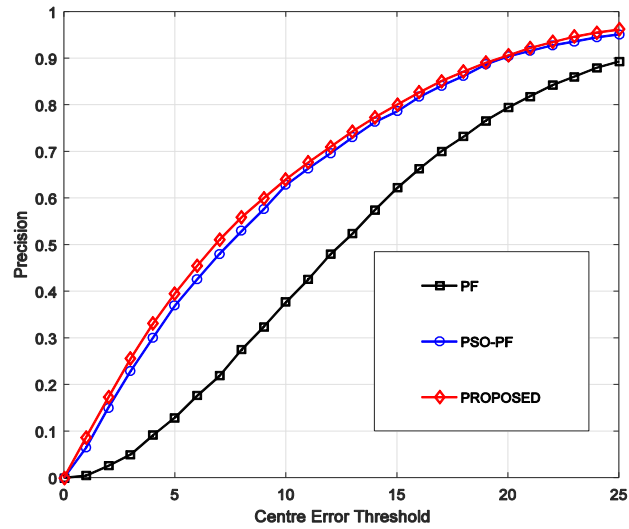

**Fig. 6.** Precision versus Center Error Threshold for dataset OneStopNoEnter2cor.

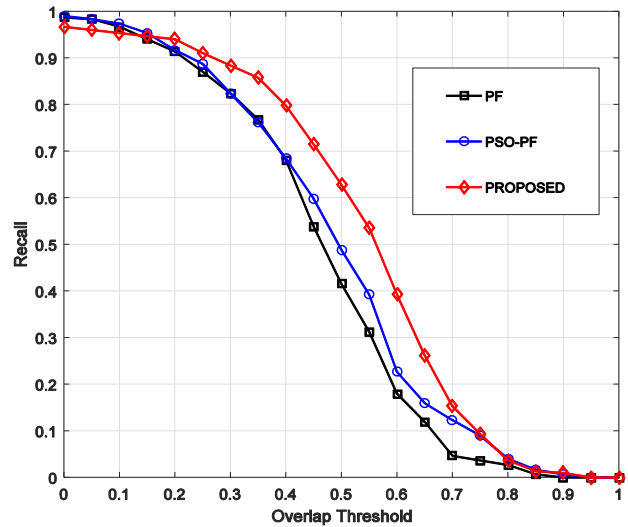

**Fig. 7.** Recall versus Overlap Threshold for dataset OneStopNoEnter2cor.

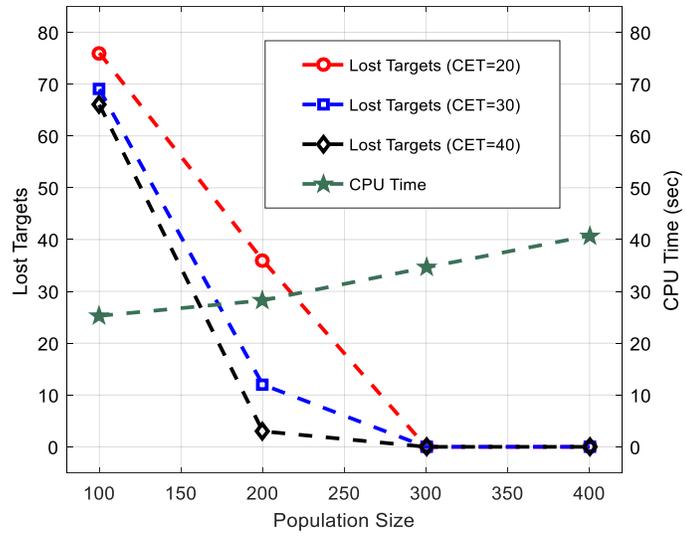

**Fig. 8**. Performance of AWQPSO under varying population sizes for dataset OneStopNoEnter2cor.

*6.3. Results for Benchmark Problem 2: aerobatics_1*

| Frame | PF | PSO-PF | AWQPSO-PF |
|---|---|---|---|
| 324 | | | |
| 432 | | | |
| 513 | | | |
| 540 | | | |
| 597 | | | |

**Fig. 9.** Tracking results for aerobatics_1.

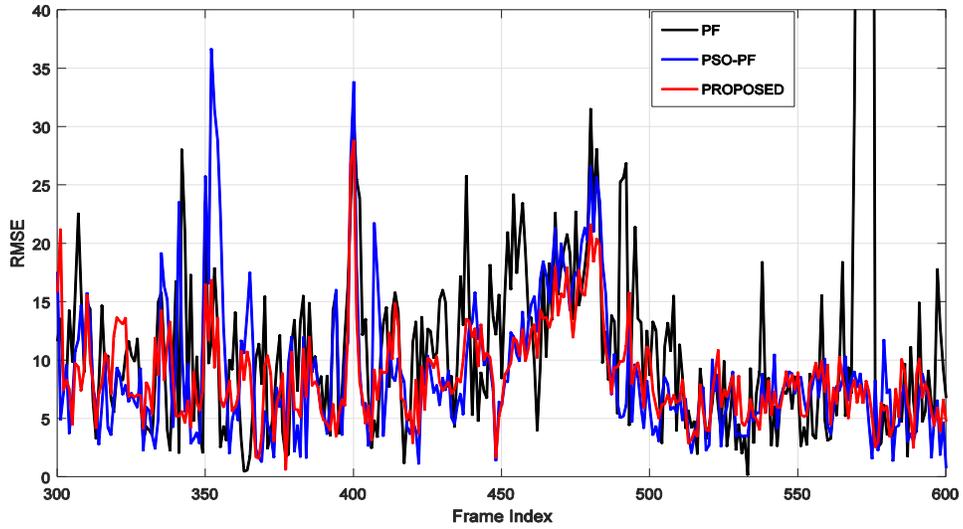

**Fig. 10.** Evolution of RMSE for aerobatics_1 (301 frames).

**Table 4**
Performance comparison of the three trackers for aerobatics_1.

| Dataset | Algorithm | FPS | Lost Targets | |
| --- | --- | --- | --- | --- |
| | | | CET=20 | CET=30 |
| aerobatics_1 | PF | **14.34±0.2016** | 32/301 | 9/301 |
| | PSO-PF | 5.40±0.4783 | 18/301 | 3/301 |
| | AWQPSO-PF | 5.79±0.3158 | **5/301** | **0/301** |

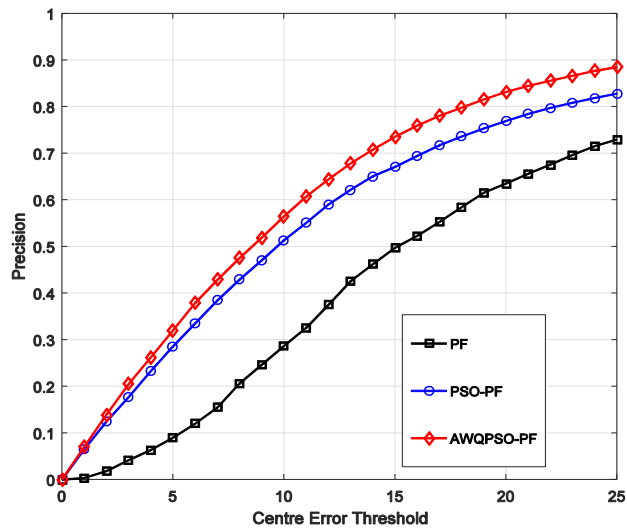

**Fig. 11.** Precision versus Center Error Threshold for dataset aerobatics_1.

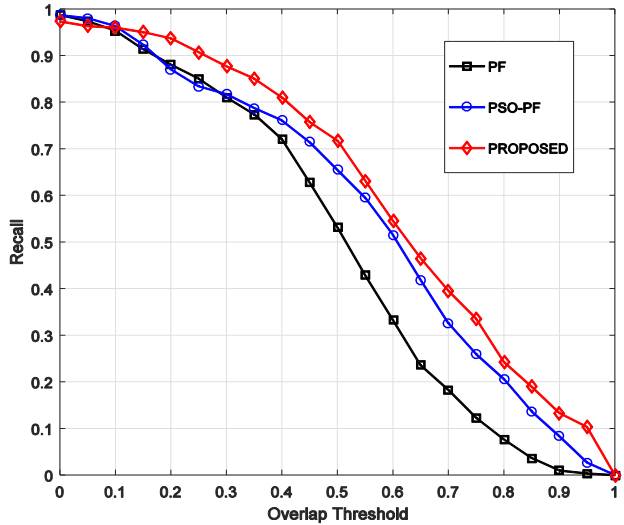

**Fig. 12.** Recall versus Overlap Threshold for dataset aerobatics_1.

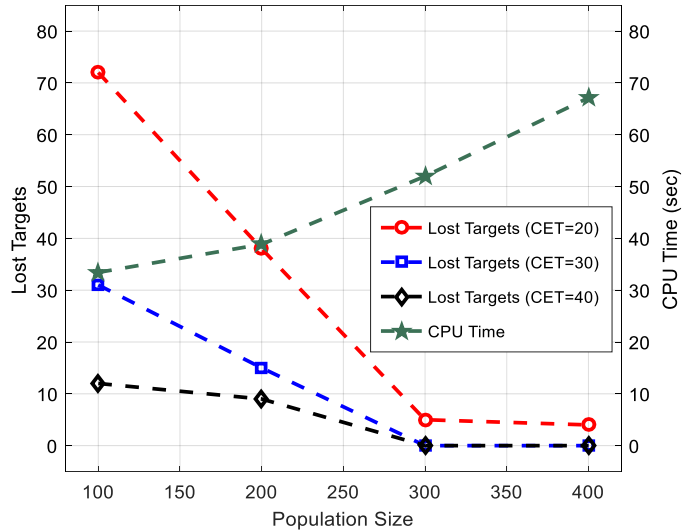

**Fig. 13.** Performance of AWQPSO under varying population sizes for dataset aerobatics_1.

## 7. Analysis of experimental results

There is an increase in FPS by 29.51% and 7.22% in case of OneStopNoEnter2cor and aerobatics_1 using AWQPSO over PSO in the Particle Filtering framework. The precision plot for OneStopNoEnter2cor suggests at least 80% of frames pass the RMSE threshold of 15 for both AWQPSO and PSO while that for aerobatics_1 suggests the same percentage of frames pass the RMSE thresholds of 18 and 23 for AWQPSO and PSO. There are 13% and 5% increases in number of frames with a 50% overlap between ground truth and tracker bounding boxes when using AWQPSO as compared to PSO for OneStopNoEnter2cor and aerobatics_1 respectively. In Frames 1075 through 1091 of OneStopNoEnter2cor, the PF tracker is distracted by mistaking local objects as the target, whereas PSO-PF and AWQPSO-PF maintain tracking the target viz. a human subject walking down the corridor clad in red clothing successfully with RMSE<10. Additionally, in aerobatics_1 for Frames 566 to 575 and 594 to 600, PF loses track of the target aircraft due to abrupt motion coupled with scale change, however PSO and AWQPSO trackers perform efficiently. In both the periods though, the proposed AWQPSO-PF tracker has a lower RMSE than the PSO-PF tracker.

Table 3 lists the results of performance parameters for the OneStopNoEnter2cor sequence using the different techniques. Although experimental results suggest that the AWQPSO-PF approach tracks the target with the least net error as compared to PF and PSO-PF, it takes at least twice as much time to process the same number of frames as the Particle Filter does. The number of lost targets for Centre Error Threshold of 20 and 30 are least in AWQPSO-PF and its RMSE is less than 20 in each of the 301 frames of the subsequence, whereas PF and PSO-PF fail to confine the RMSE to under 20 in all frames. The number of correctly tracked frames (no lost targets) given a RMSE threshold of 20 rose by 1.328% and 13.289% using the proposed approach over PSO and PF respectively. While the AWQPSO-PF and PSO-PF approaches reported same number of correctly tracked frames for RMSE threshold of 30, there was an increase of 9.302% noticed with regard to the PF performance for AWQPSO-PF.

Results from Table 4 indicate AWQPSO-PF has a much tighter bounding box around the target in each frame when compared to the other methods. For instance, the number of frames in the subsequence where the swarm RMSE is less than or equal to 20 is 296 and 283 in case of AWQPSO-PF and PSO-PF respectively – an improvement of 4.318%. Similarly, the concerned number of frames are 298 and 301 for swarm RMSE less than or equal to 30 meaning an improvement of 0.996% using AWQPSO-PF over PSO-PF. The proposed approach reported 8.970% and 2.990% increase in said number of frames for RMSE bounds of 20 and 30 against the standard PF for the AWQPSO-PF tracker.

## 8. Conclusion and future work

The present study has presented and tested an evolutionary Particle Filter which makes use of an Annealed - Quantum-behaved Particle Swarm Optimization with a weighted Mean Best operator. The better global search ability of the fitness weighted QPSO along with the probabilistic rejection of inferior solutions using Metropolis Criterion makes the proposed metaheuristic well suited for avoiding local minima in the tracking search space. This preserves the diversity of the posterior population and alleviates the sample impoverishment issue to an extent better than the competing Particle Swarm Optimization based Particle Filter and the standard Particle Filter. This is evidenced by the experimental results obtained in Tables 3 and 4 as well as by the metrics in Figures 6, 7, 11 and 12. In addition to this, a motion model that looks back three steps in memory is adopted to smooth out sudden changes in velocity of the target. The proposed algorithm is tested using two sequences and is seen to outperform its competitors in both, yielding better RMSE across majority of frames as well as greater area under the curve for both the Precision versus Centre Error Threshold and Recall versus Overlap Threshold metrics. It is observed that the computational load for the AWQPSO-PF method is lower than the PSO-PF, albeit both being significantly slower than the standard PF tracker. This is because of the lesser number of within-frame iterations required by AWQPSO to reach the convergence threshold. However, given the large number of particles used in all the methods and the large within-frame cutoff iteration of 50, the setup is not suitable for real time operation without a reduction in population size and number of in-frame iterations or a parallelized implementation.

The observation model may be modified to accommodate a multi cue likelihood function requiring a multi-objective optimization approach thus effecting a better representation of the target. Additionally, the current AWQPSO-PF tracker model can be extended to track multiple targets with a focus on occlusion handling and evasion of stagnation in local minima over a large number of datasets. Importantly enough, the speedup through parallel computation of particle trajectories in the dynamic state transition section and the subsequent metaheuristic optimization module may lead to a significant increase in FPS. As with existing swarm optimization inspired tracking models such as the Cuckoo Search inspired PF tracker in [9], the QPSO-PF tracker in [7], the Cellular QPSO-PF tracker in [8] and other recent ones [10-11], the current metaheuristic too is susceptible to performance degradation due to incorrect parametric tuning, necessitating a thorough characterization of the operating ranges of its system variables to guarantee convergent behavior.

### Acknowledgements

This work was made possible by the financial and computing support by the Vanderbilt University Department of EECS.

### Conflict of Interest

The authors declare that there is no conflict of interest regarding the publication of this article.